\documentclass[graybox]{svmult}

\usepackage{mathptmx}
\usepackage{helvet}
\usepackage{courier}
\usepackage{type1cm} 
\usepackage{makeidx} 
\usepackage{graphicx} 
\usepackage{multicol} 
\usepackage[bottom]{footmisc}
\usepackage{subfigure}
\usepackage{microtype}
\usepackage{amsmath}
\newcommand{\argmax}{\operatornamewithlimits{argmax}}

\begin{document}

\title*{Co-Clustering Network-Constrained Trajectory Data}

\author{Mohamed K. El Mahrsi, Romain Guigour\`es, Fabrice Rossi, and Marc Boull\'e}
\institute{Mohamed K. El Mahrsi \at T\'el\'ecom ParisTech - D\'epartement Informatique et R\'eseaux, 46 Rue Barrault 75634 Paris CEDEX 13 France, \email{khalil.mahrsi@telecom-paristech.fr}
\and 
Romain Guigour\`es \at Orange Labs, 2 Avenue Pierre Marzin 22300 Lannion France,
\email{romain.guigoures@orange.com}
\and Fabrice Rossi \at \'Equipe SAMM EA 4543, Universit\'e Paris I Panth\'eon-Sorbonne, 90 Rue de Tolbiac 75634 Paris CEDEX 13 France, \email{fabrice.rossi@univ-paris1.fr}
\and Marc Boull\'e \at Orange Labs, 2 Avenue Pierre Marzin 22300 Lannion France, \email{marc.boulle@orange.com}
}
\maketitle

\abstract*{Recently, clustering moving object trajectories kept gaining interest from both the data mining and machine learning communities. This problem, however, was studied mainly and extensively in the setting where moving objects can move freely on the euclidean space. In this paper, we study the problem of clustering trajectories of vehicles whose movement is restricted by the underlying road network. We model relations between these trajectories and road segments as a bipartite graph and we try to cluster its vertices. We demonstrate our approaches on synthetic data and show how it could be useful in inferring knowledge about the flow dynamics and the behavior of the drivers using the road network.}

\abstract{Recently, clustering moving object trajectories kept gaining interest from both the data mining and machine learning communities. This problem, however, was studied mainly and extensively in the setting where moving objects can move freely on the euclidean space. In this paper, we study the problem of clustering trajectories of vehicles whose movement is restricted by the underlying road network. We model relations between these trajectories and road segments as a bipartite graph and we try to cluster its vertices. We demonstrate our approaches on synthetic data and show how it could be useful in inferring knowledge about the flow dynamics and the behavior of the drivers using the road network.}

\section{Introduction}

Monitoring traffic on road networks is generally handled using dedicated sensors that provide estimations of the number of vehicles traversing the road portion on which they are deployed. Due to their prohibitive installation and maintenance costs, the deployment of these sensors is mainly limited to the primary road network (i.e. highways and main arteries). Consequently, the road network's state reported using this kind of solutions is partial and incomplete which complicates the application of data mining tasks that aim to extract meaningful knowledge about flow dynamics and mobility patterns.

Thanks to the advances in the fields of telecommunication and geo-positioning, an alternative approach may consist in taking advantage of GPS logs collected on moving objects that are equipped with ad hoc devices (such as smartphones). These logs can be acquired through dedicated data acquisition campaigns (using probing vehicles, buses, taxis, etc.), through crowdsourcing mechanisms in which users contribute their own trajectories, etc. Trajectory data can thus be harvested on a large scale which helps provide a better coverage of the road network compared to sensor data.

Clustering is a widely used technique in exploratory data analysis. Given a set of observations, cluster analysis consists in partitioning these observations into groups (called clusters) in such fashion that objects belonging to the same group are more similar to each other (w.r.t. a given criterion) than to objects from other groups. Most prior work on trajectory clustering focused on the case of moving objects evolving freely in the euclidean space \cite{Kalnis_2005, Benkert_2006, Lee_2007, Jeung_2008a}. Often in real applications, however, moving objects must comply with the existence of an underlying network (for instance, vehicles evolve in the road network, airplanes must remain in invisible but well defined air corridors, etc.). The topological constraints imposed by this network play a key role in determining the similarity between trajectories and should logically be accounted for during the clustering process. Clustering network-constrained trajectories gained in popularity only recently with the publication of work such as \cite{Kharrat_2008, Liu_2008, Roh_2010}, etc.

The insightful idea of using a graph-based approach to cluster trajectory data was first used in \cite{Guo_2010}. We built upon the premisses of this idea in \cite{Mahrsi_2012a} where we used a graph representation to model the similarity relationships between trajectories and clustered this similarity graph to extract clusters of trajectories that exhibit similar mobility patterns. This approach was extended in \cite{Mahrsi_2012c} and used to regroup similar road segments that can eventually be used to further enhance the interpretability of trajectory clusters. In the present work, we retain this idea of using a graph representation as we model the interactions between trajectories and road segments using a bipartite graph and we study two different approaches to clustering its vertices.

The remainder of this paper is organized as follows. Our data model and proposed approaches are presented in Sect.~\ref{sec:ClusteringApproaches}. Section~\ref{sec:ExperimentalStudy} illustrates our experimental study where we demonstrate our propositions' capacity to highlight and discover interesting trajectory and road segment clusters. Related work is briefly discussed in Sect.~\ref{sec:RelatedWork}. Finally, conclusions and future work are presented in Sect.~\ref{sec:Conclusions}.

\section{Clustering Approaches}
\label{sec:ClusteringApproaches}

In the network-constrained case, trajectories are often modeled using a symbolic data model \cite{Kharrat_2008, Lou_2009, Roh_2010}. Each trajectory $T$ is represented as a series of succeeeding road segments (this is done by applying a map-matching algorithm, such as \cite{Lou_2009}, to the original GPS logs). Therefore, two entities are eligible for applying clustering techniques: (i) trajectories, and (ii) road segments.

\begin{figure}[h]
\includegraphics[scale=.28]{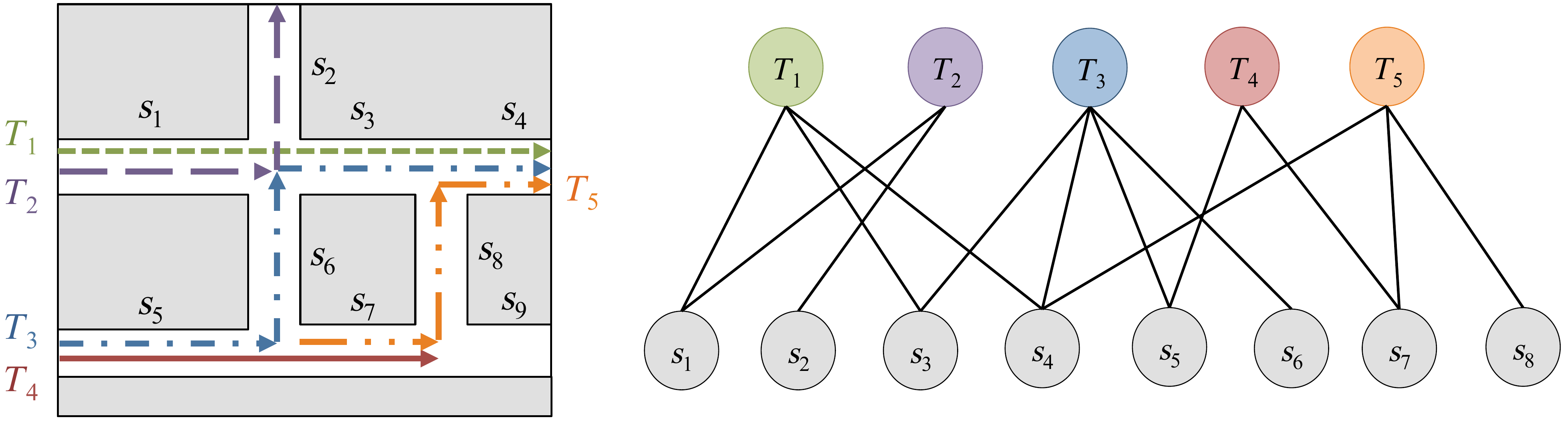}
\caption{A bipartite graph is used to model interactions between the trajectories and the road network's segments. Each trajectory and each road segment is represented as a vertex in the graph. Edges are created between each trajectory and the set of road segments it visited.}
\label{fig:bigraph}
\end{figure}

We model the data as a bipartite graph $\mathcal{G} = (\mathcal{T}, \mathcal{S}, \mathcal{E})$. $\mathcal{T} = \left\{T_1, T_2, ..., T_n \right\}$ is the dataset of trajectories, $\mathcal{S} = \left\{s_1, s_2, ..., s_m \right\}$ is the set of all the road segments composing the road network that registered at least one traversal, and $\mathcal{E}$ is the set of edges modeling interactions between trajectories and road segments (i.e. an edge $e$ exists between a trajectory $T$ and a road segment $s$ if and only if $T$ visited $s$ at least once). This representation is illustrated in Fig.\ref{fig:bigraph} depicting five trajectories $T_1$, $T_2$, $T_3$, $T_4$, and $T_5$ interacting with eight road segments and the corresponding bipartite graph.

We will first attempt to project the bipartite graph $\mathcal{G}$ on both its trajectory vertices $\mathcal{T}$ and road segment vertices $\mathcal{S}$ and study clustering the resulting graphs separately (Sect.~\ref{sec:ClusteringProjectedGraphs}). Secondly, we will process $\mathcal{G}$ directly using a co-clustering approach (Sect.~\ref{sec:CoClusteringBipartiteGraph}).

\subsection{Clustering the Projected Trajectory and Segment Graphs}
\label{sec:ClusteringProjectedGraphs}

\begin{figure}[h]
\center
\includegraphics[scale=.3]{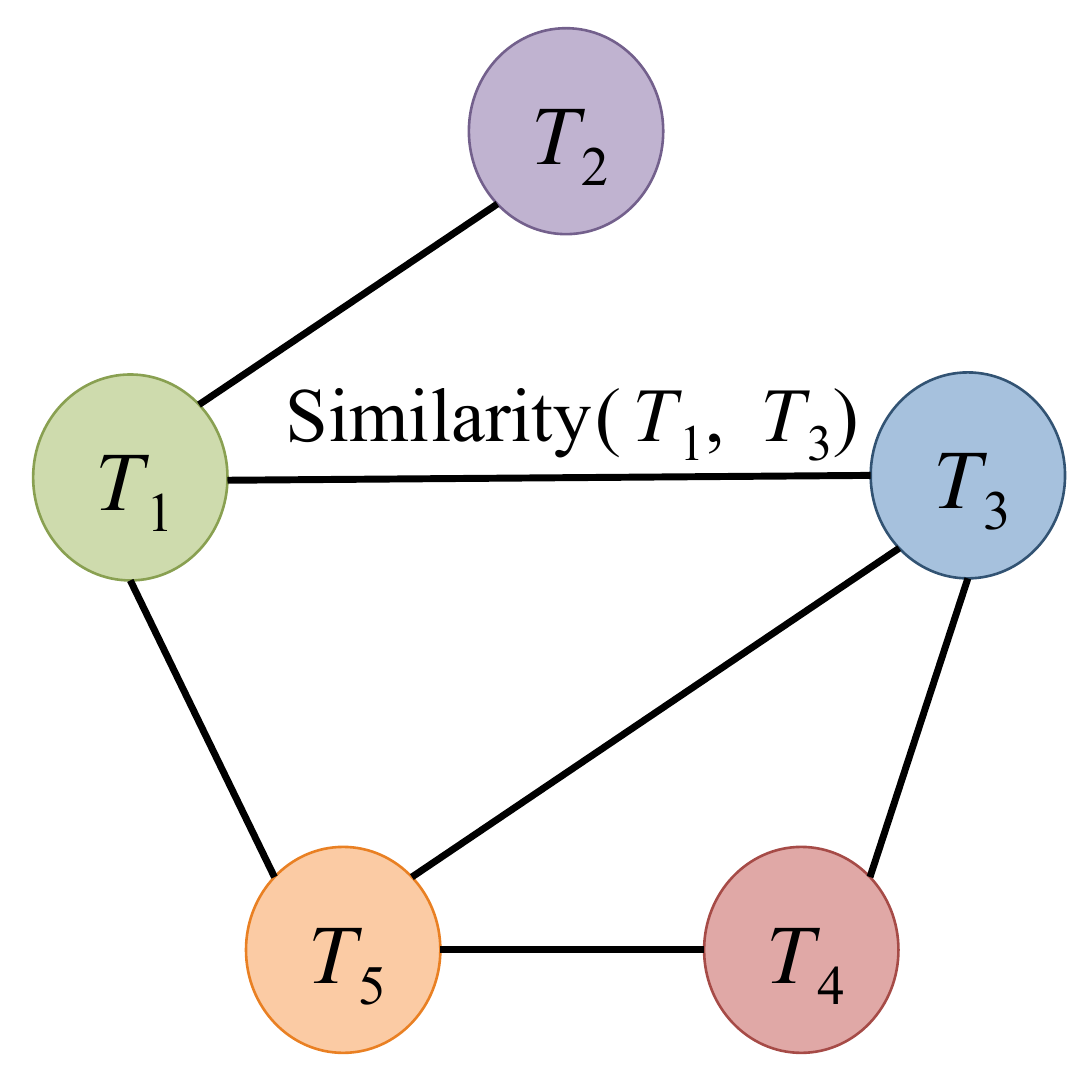}
\caption{The trajectory similarity graph resulting from the projection of the bipartite graph depicted in Fig.~\ref{fig:bigraph} on its trajectory vertices. Here, each trajectory is represented as a vertex and weighted edges inter-connect trajectories based on their similarity.}
\label{fig:trajsimgraph}
\end{figure}

Our bipartite graph $\mathcal{G}$ is composed of two types of vertices, trajectory vertices and road segment vertices. We can project $\mathcal{G}$ on the set of trajectory vertices $\mathcal{T}$ which produces a new, simple graph $\mathcal{G}_\mathcal{T} = (\mathcal{T}, \mathcal{E}_\mathcal{T}, \mathcal{W}_\mathcal{T})$ that represents similarity relationships between trajectories. In this setting, $\mathcal{T}$ stands for vertices representing trajectories, $\mathcal{E}_\mathcal{T}$ are edges indicating the presence of similarities between pairs of trajectories (an edge $e_{\left<T_i, T_j\right>}$ exists between two trajectories $T_i$ and $T_j$ if they share at least one common road segment), and $\mathcal{W}_\mathcal{T}$ are weights assigned to edges based on the strength of the similarity between trajectories they connect. An example of a projected trajectory similarity graph is depicted in Fig.~\ref{fig:trajsimgraph}.

The most basic weighting strategy is to assign to each edge $e_{\left<T_i, T_j\right>}$ a weight $\omega_{\left<T_i, T_j\right>}$ that is equal to the count of common road segments between the two trajectories $T_i$ and $T_j$. However, the main drawback of this approach is that it completely neglects the spatial properties of road segments (for instance, a very short road segment and a lengthy one have equal contributions to the similarity). Therefore, we proposed a more sophisticated and spatially-aware weighting strategy in \cite{Mahrsi_2012a}. It is this strategy that we will continue to use here.

For each road segment $s$ visited by a trajectory $T$, we calculate its contribution (w.r.t. this trajectory) based not only on its spatial length but also on the frequency of its appearance in the dataset. This contribution is basically an adaptation of tf-idf (term frequency - inverse document frequency) weighting widely used in information retrieval, modified to account for spatiality. The contribution $w_{s,T}$ of segment $s$ to trajectory $T$ is calculated according to Eq.~(\ref{eq:SegmentWeight}).

\begin{equation}
\label{eq:SegmentWeight}
w_{s,T} = \frac{n_{s, T} \cdot \mbox{length}(s)}{\sum_{s' \in T} n_{s', T} \cdot \mbox{length}(s')} \cdot \log\frac{| \mathcal{T} |}{| \{T_i : s \in T_i \} |}~.
\end{equation}

$n_{s, T}$ is the number of times the trajectory $T$ visited the road segment $s$ (usually equal to 1), $\mbox{length}(s)$ is the spatial length of the segment. $| \mathcal{T} |$, the total number of trajectories in $\mathcal{T}$, and $| \{T_i : s \in T_i \} |$, the number of trajectories that visited $s$. The second term in $w_{s,T}$ is used to penalize frequently-traveled road segments (with the intuition that the more a segment is traveled, the less it is relevant w.r.t. similarity evaluation and vice versa).

We evaluate the similarity between pairs of trajectories using a cosine similarity and we assign the weights in $\mathcal{G}_\mathcal{T}$ accordingly (\ref{eq:TrajectoryCosineSimilarity}):

\begin{equation}
\label{eq:TrajectoryCosineSimilarity}
\omega_{\left<T_i, T_j\right>} = \frac{\sum_{s \in \mathcal{S}} w_{s,T_i} \cdot w_{s,T_j} }{\sqrt{\sum_{s \in \mathcal{S}} w_{s,T_i}^2} \cdot \sqrt{\sum_{s \in \mathcal{S}} w_{s,T_j}^2}}~.
\end{equation}

By analogy, we can project the bipartite graph $\mathcal{G}$ on the set of road segment vertices $\mathcal{S}$ in which case we obtain a segment similarity graph $\mathcal{G}_\mathcal{S} = (\mathcal{S}, \mathcal{E}_\mathcal{S}, \mathcal{W}_\mathcal{S})$. In this graph, a similarity edge $e_{\left<s_i, s_j\right>}$ indicates that at least one trajectory visited both road segments $s_i$ and $s_j$. Here again, it is totally feasible to assign edge weights based solely on the count of common trajectories, but instead we define a weighting technique based on trajectory relevance \cite{Mahrsi_2012c} similarly to what we did earlier when processing trajectories. The similarity between two road segments $s_i$ and $s_j$ is expressed as follows (\ref{eq:SegmentCosineSimilarity}):

\begin{equation}
\label{eq:SegmentCosineSimilarity}
\omega_{\left<s_i,s_j\right>} = \frac{\sum_{T \in \mathcal{T}} w_{T,s_i} \cdot w_{T,s_j}}{\sqrt{\sum_{T \in \mathcal{T}} w_{T,s_i}^2} \cdot \sqrt{\sum_{T \in \mathcal{T}} w_{T,s_j}^2}}~.
\end{equation}

With :
\begin{equation}
	w_{T, s} = \frac{n_{s,T}}{\sum_{T' \in \mathcal{T}} n_{s,T'}} \cdot \log\frac{|\mathcal{S}|}{|s' \in \mathcal{S}: s' \in T|}~.
\end{equation}

The first part of $w_{T, s}$ evaluates the ``contribution'' (or importance) of trajectory $T$ to the road segment $s$ by calculating the ratio between the number of visits $n_{s,T}$ made by $T$ to $s$ and the number of visits $s$ received from all the trajectories in $\mathcal{T}$. The second parts evaluates the overall relevance of $T$ based on comparing the number of different segments it visited $|s' \in \mathcal{S}: s' \in T|$ to the total number of road segments $|\mathcal{S}|$.

We propose to cluster the projected trajectory and segment graphs separately in order to discover trajectory clusters on one side and road segment clusters on the other. To do so, we chose to apply modularity-based community detection using an algorithm that implements the directives described in \cite{Noack_2009}. This choice is mainly motivated by the fact that vertices in such similarity graphs are expected to have high degrees and modularity-based clustering is reputed to outshine other approaches in such settings. Nevertheless, we do not exclude the use of other graph clustering algorithms (e.g. spectral clustering \cite{Meila_2000}) if these can yield better results.

The used algorithm produces a hierarchy of nested (trajectory or segment) clusters that are suitable for multi-level exploration where the user can start by inspecting a few, coarse clusters in order to quickly understand the general motion trends then proceed to exploring clusters of interest with higher levels of detail by means of successive refinements. Also, once the trajectory and segment partitions are found, they can either be analyzed separately or cross-compared and interpreted based on each other.

Given a dataset of $n$ trajectories that travelled on a road network composed of $m$ segments, the time complexity for clustering the trajectory graph is theoretically in $O(n^3)$ whereas clustering road segments is done in $O(m^3)$ \cite{Noack_2009}.

\subsection{Direct Co-Clustering of the Bipartite Graph}
\label{sec:CoClusteringBipartiteGraph}

We now propose to study clustering the bipartite graph $\mathcal{G}$ directly. To achieve this end, we apply a co-clustering approach to the graph's adjacency matrix. In the adjacency matrix, trajectories are represented in the rows whilst road segments are represented on the columns. The intersection of row $i$ with column $j$ indicates the number of times trajectory $T_i$ visited the road segment $s_j$ ($1 \leq i \leq n$ and $1 \leq j \leq m$). Co-clustering works by rearranging the rows and columns of the adjacency matrix in order to highlight blocs that have homogeneous density. These blocs are then used to derive two partitions simultaneously (one partition for trajectories and the other for road segments in our case). A co-clustering structure, that we refer to as $\mathcal{M}$ hereafter, is usually defined through a set of modeling parameters. Ours are described in Table~\ref{tab:notations}. 

\begin{table}[ht]
	\caption{Notations.}
	\label{tab:notations}
		\begin{tabular}{|p{0.45\linewidth}|p{0.45\linewidth}|}
		\hline
		\textbf{Bipartite graph $\mathcal{G}$} &\textbf{Co-clustering model $\mathcal{M}$ }\\
		\hline
		$\mathcal{T}$ : set of trajectories & $C_{\mathcal{T}}$ : set of trajectory clusters \\
		$\mathcal{S}$ : set of road segments & $C_{\mathcal{S}}$ : set of road segment clusters \\
		$\mathcal{E}=\mathcal{T} \cap \mathcal{S}$ : set of traversals of road segments in $\mathcal{S}$ by trajectories in $\mathcal{T}$ & $C_{\mathcal{E}}=C_{\mathcal{T}} \cap C_{\mathcal{S}}$ : co-clusters of trajectory and road segments\\
		\hline			
		\end{tabular}
\end{table}

The objective of co-clustering algorithms is to infer the best partition of the bipartite graph. By applying such approaches, trajectories are regrouped if they travel along common road segments and, vice versa, road segments are clustered together if they are visited by the same trajectories. The main advantage of these techniques is that they do not require preprocessing nor do they require the definition of an ``artificial'' similarity between trajectories or between segments. Nonetheless, they do present the drawback of being computationally expensive.

We opt for the MODL \cite{Boulle_2011} approach to conduct the co-clustering of $\mathcal{G}$. We made this choice because this approach (i) is non-parametric and does not require user intervention or fine-tuning, (ii) is easily scalable and can, consequently, be used to analyze large datasets, and (iii) was already and successfully applied to geo-tagged data \cite{Guigoures_2012}. In MODL, a quality criterion is defined according to a Maximum A Posteriori (MAP) approach (\ref{eq:maximumaposteriori}):

\begin{equation}
\label{eq:maximumaposteriori}
\mathcal{M}^*=\argmax_{\mathcal{M}} P(\mathcal{M}) P(\mathcal{D} | \mathcal{M})~.
\end{equation}

First, an a priori probability $P(\mathcal{M})$ is defined based on the data (denoted $\mathcal{D}$). This probability tries to characterize each of the modeling parameters of the model $\mathcal{M}$ by assigning to each one of them a penalty (which corresponds to their minimal coding length, calculated based on descriptive statistics of the data). Next, the likelihood of the data given the data model $P(\mathcal{D}|\mathcal{M})$ is defined. The likelihood measures the cost
of re-encoding $\mathcal{D}$ with the parameters of $\mathcal{M}$. Consequently, the most likely co-clustering model is the one that is most faithful to the original data (in other terms, the likelihood tends to favor relevant and informative structures). Retrieving the best co-clustering (i.e. the one optimizing the global criterion $\mathcal{M}^*$) consists in realizing the best trade-off between conciseness and accuracy.

Since co-clustering problems are often $NP$-complete, the clustering is conducted using an agglomerative greedy heuristic. Initially, the trivial, most refined model is considered (this model contains only one trajectory and one road segment per cluster). Then, all cluster merging operations are evaluated and the best merge is applied (if it results in a decrease of the quality criterion). Once no more merging operations are possible, the result of the heuristic is refined using a post-optimization step in which some elements swap their cluster memberships. The whole process is encapsulated within a VNS (Variable Neighborhood Search, \cite{Hansen_2001}) metaheuristic that restarts the algorithm several times with different random cluster initializations. Full details and a thorough evaluation of the MODL approach can be retrieved in \cite{Boulle_2011}.

MODL has a complexity of $O(|\mathcal{E}| \sqrt{|\mathcal{E}|} \log(|\mathcal{E}|))$ where $\mathcal{E}$ indicates the total number of edges in the bipartite graph $\mathcal{G}$ (which, in our case, translates to the overall number of road segment traversals). This complexity, however, is only observed in the worst and very unlikely case where each trajectory in the dataset $\mathcal{T}$ visits each single road segment in the road network.

\section{Experimental Study}
\label{sec:ExperimentalStudy}

In this section, we demonstrate how the proposed approaches can be used to discover and analyze motion patterns in road networks. Our experimental setting is described in Sect.~\ref{sec:ExperimentalSetting} whereas results and their interpretation are presented in Sect.~\ref{sec:AnalysisTrajectoryClusters} and Sect.~\ref{sec:MutualClusterAnalysis}.

\subsection{Experimental Setting}
\label{sec:ExperimentalSetting}

In order to test our propositions, we use synthetic datasets of labeled trajectories. These datasets are generated intentionally to contain trajectories that are supposed to form natural clusters using the following strategy. The space covered by the road network (represented by the minimum bounding rectangle regrouping all of its vertices) is divided into a grid of equally-sized rectangular cells (or zones). For each of the clusters to be generated, a zone is selected randomly and all of its contained vertices are chosen to play the role of departure points. Similarly, a second zone (different from the first) is also selected randomly and its vertices are used as arrival points. For each trajectory to be included in the cluster, a departure (resp. arrival) vertex is drawn randomly from the set of departure (resp. arrival) vertices and the trajectory is generated as the set of road segments forming the shortest path linking the two vertices (the shortest path calculation is based on travel time and takes into account the characteristics of the visited road segments such as speed limitations, etc.). The number of trajectories in each cluster is fixed randomly in-between two minimum and maximum user-defined values. The data generation process is conceived in such fashion that interactions between clusters can occur (examples of interactions include clusters converging from different departure zones to a common arrival zone, inverted clusters where the departure zone of one cluster is the arrival zone of the other and vice versa, etc.).

\begin{figure}[ht]
\centering
	\subfigure[Class 1 (14 trajectories)]{
		\includegraphics[width=0.31\textwidth, keepaspectratio]{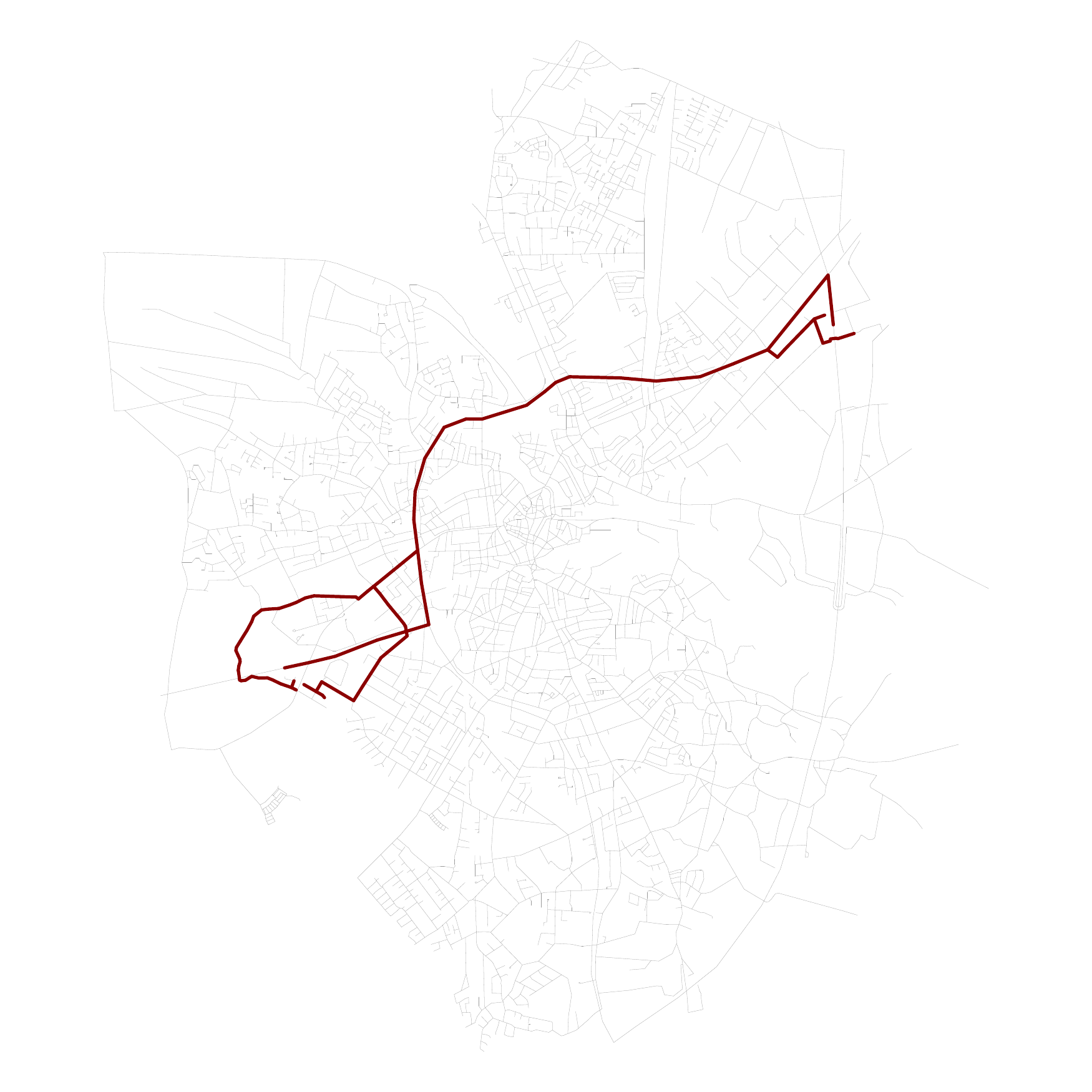}
	}
	\subfigure[Class 2 (19 trajectories)]{
		\includegraphics[width=0.31\textwidth, keepaspectratio]{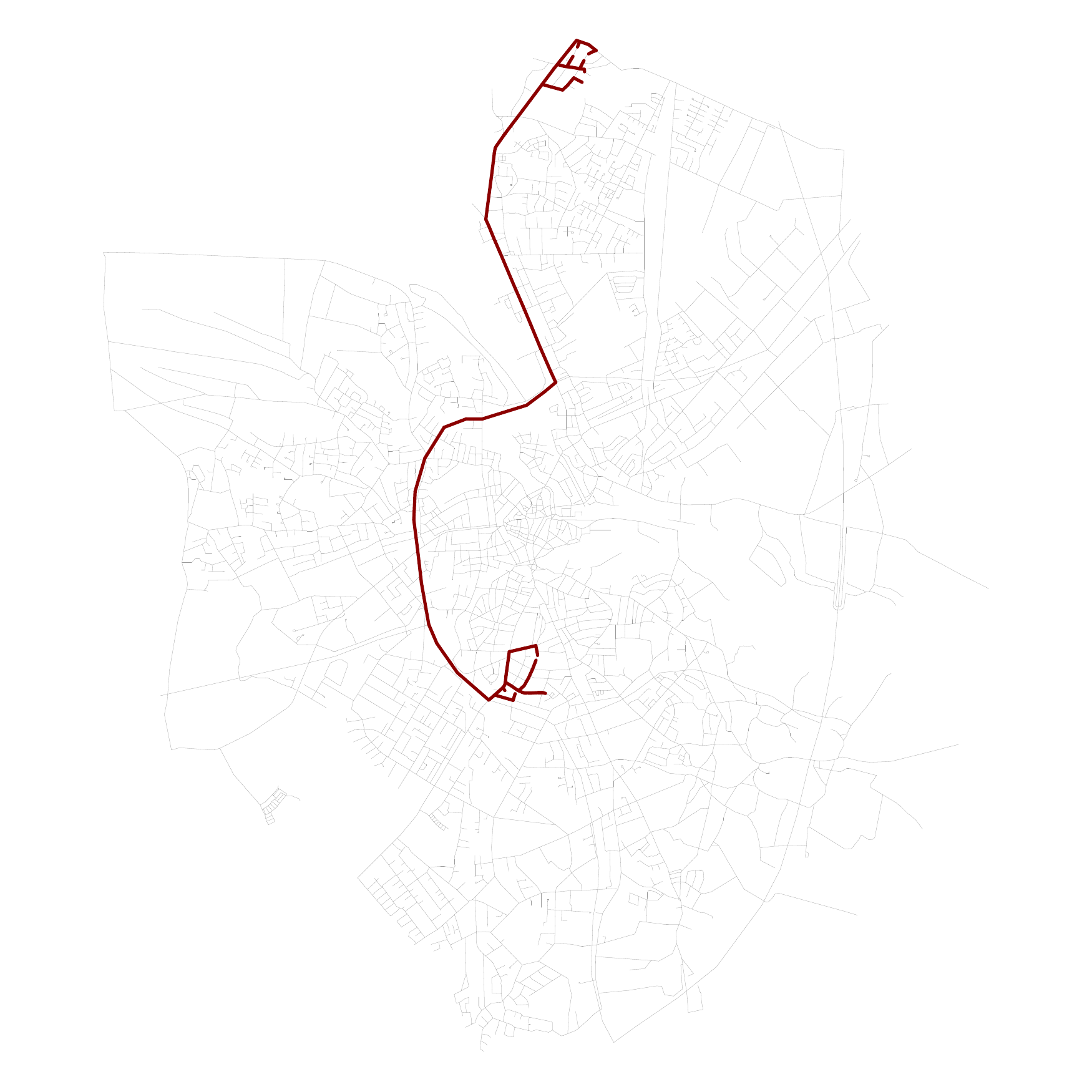}
	}
	\subfigure[Class 3 (20 trajectories)]{
		\includegraphics[width=0.31\textwidth, keepaspectratio]{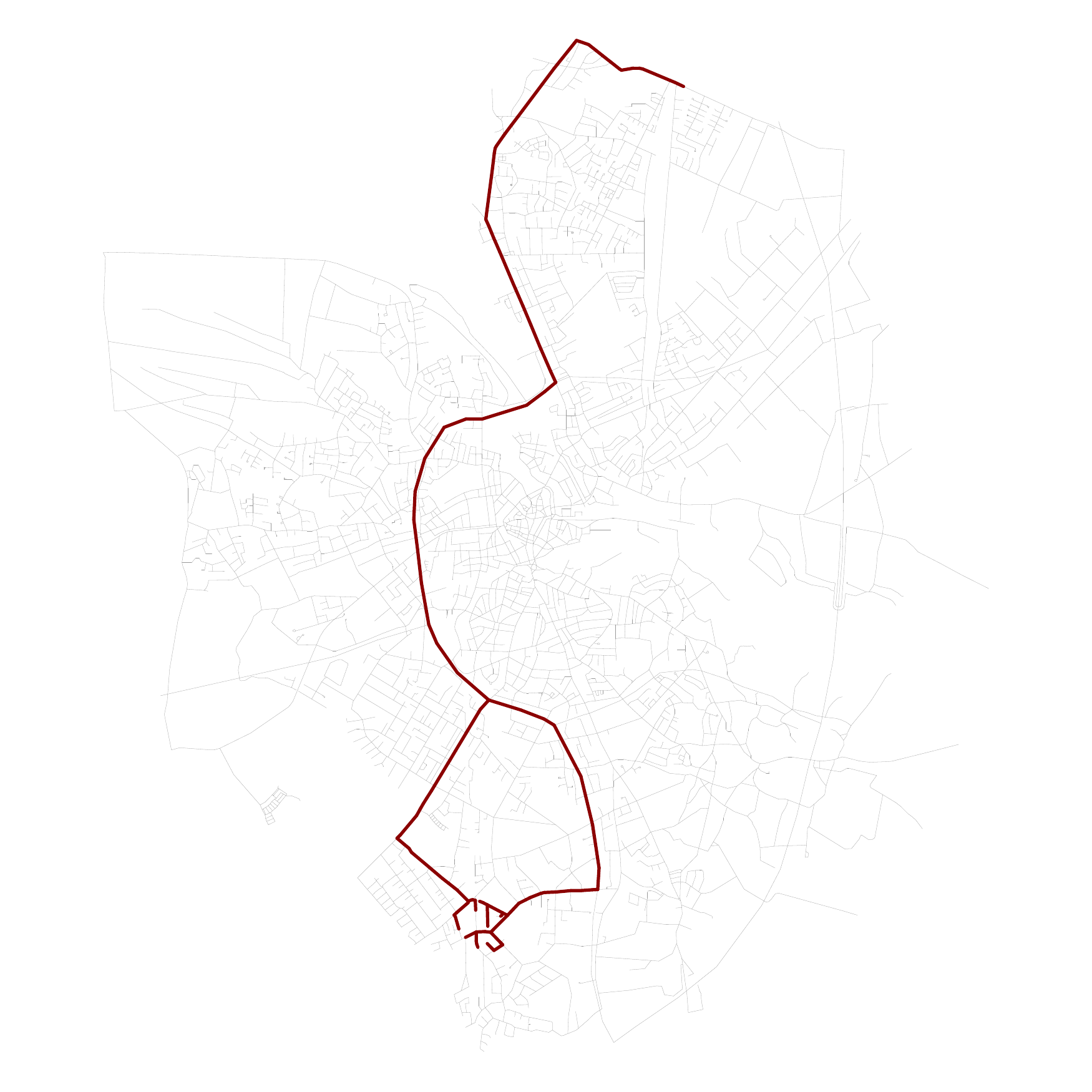}
	}
	\subfigure[Class 4 (20 trajectories)]{
		\includegraphics[width=0.31\textwidth, keepaspectratio]{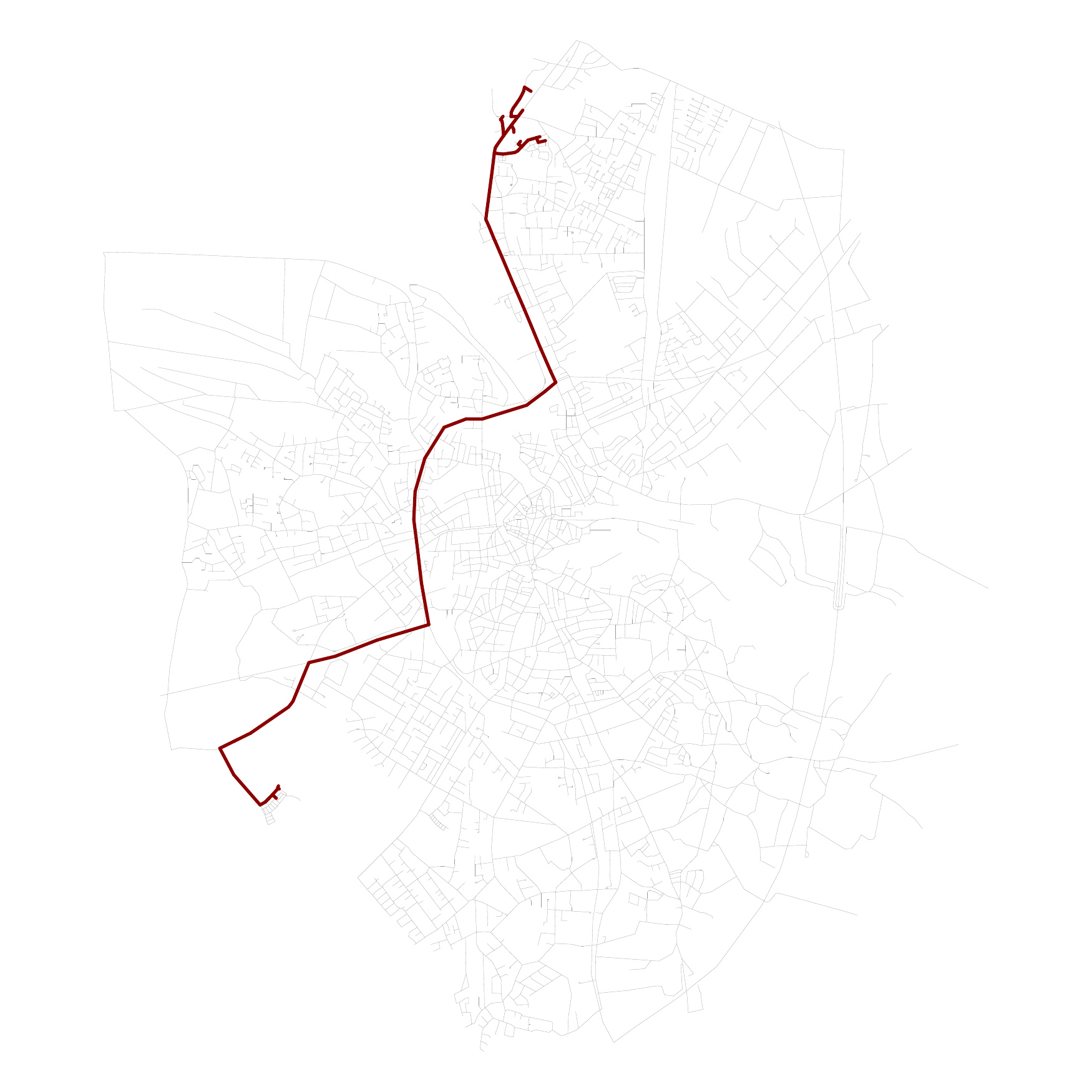}
	}
	\subfigure[Class 5 (12 trajectories)]{
		\includegraphics[width=0.31\textwidth, keepaspectratio]{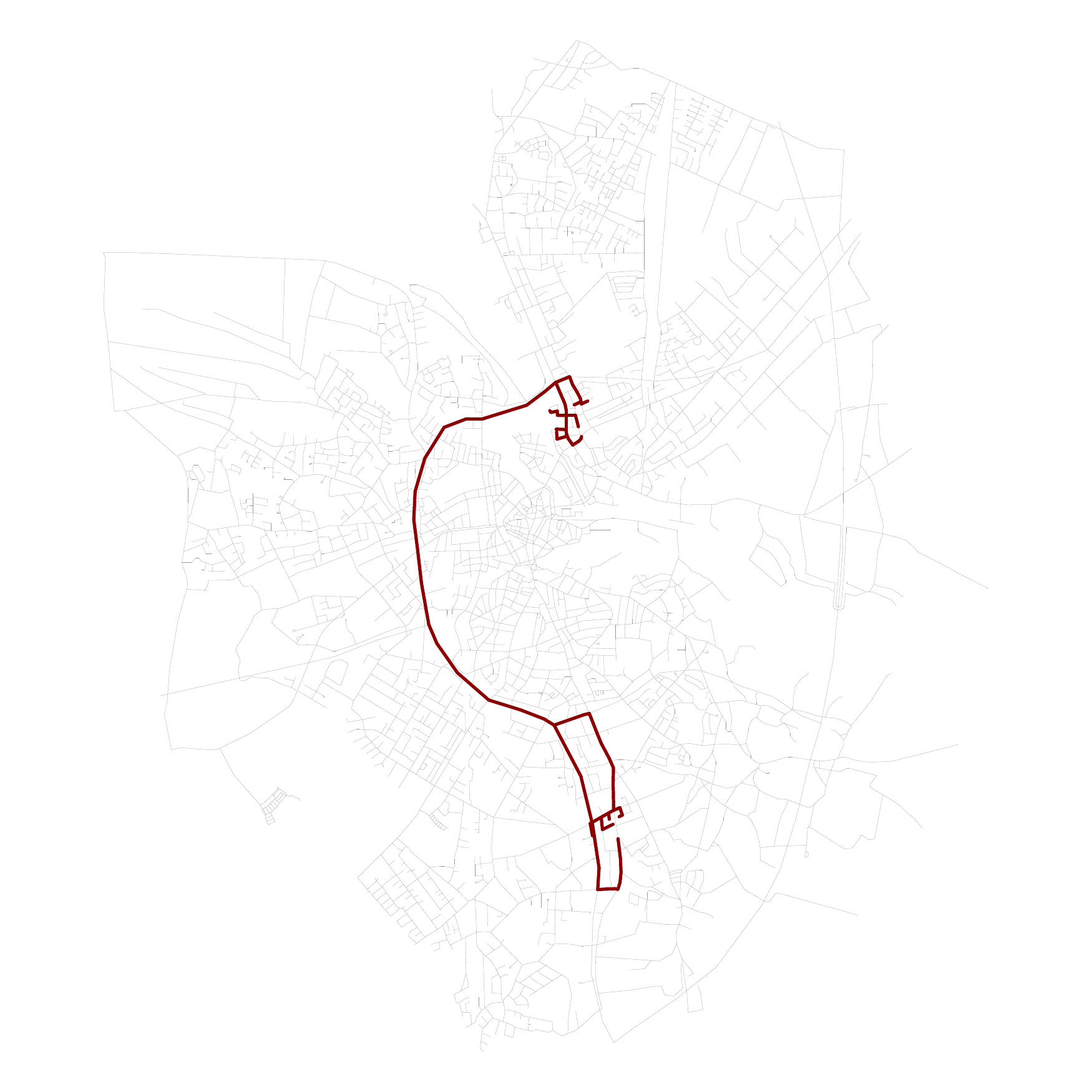}
	}
\caption{Original classes in the dataset. Some of the classes present natural interactions. For instance, classes 2 and 3 start from the same departure zone and travel together for a given portion then diverge to different arrival zones. They also interact with class 1 in the central portion of the road network.}
\label{fig:originalclasses}
\end{figure}

Since this experimental study is intended to showcase how our approaches can contribute to discovering meaningful knowledge about mobility in the road network, we make do with a case study involving a small dataset composed of 85 trajectories. These trajectories are spread across five distinct clusters (depicted in Fig.~\ref{fig:originalclasses}) and visited 485 road segments in total. We designate these original clusters as ``classes" hereafter in order to distinguish them from those that will be retrieved using the clustering algorithms. The dataset is generated using the Oldenburg road network's graph (originally provided with the Brinkhoff generator\cite{Brinkhoff_2002}) which is composed of 6105 vertices (i.e. road intersections) and 14070 directed edges (i.e. road segments).

\subsection{Analysis of Trajectory Clusters}
\label{sec:AnalysisTrajectoryClusters}

Applying modularity-based clustering to the projected trajectory similarity graph produces a partition containing three clusters (in contrast with the original five classes). This is mainly due to the fact that some classes interact considerably. Their interactions were not visible when labeling the classes individually during the data generation process. The clustering algorithm, however, was able to detect these interactions and regroup the trajectories accordingly.

\begin{figure}[ht]
\centering
	\subfigure[Modularity]{
		\includegraphics[width=0.48\textwidth, keepaspectratio]{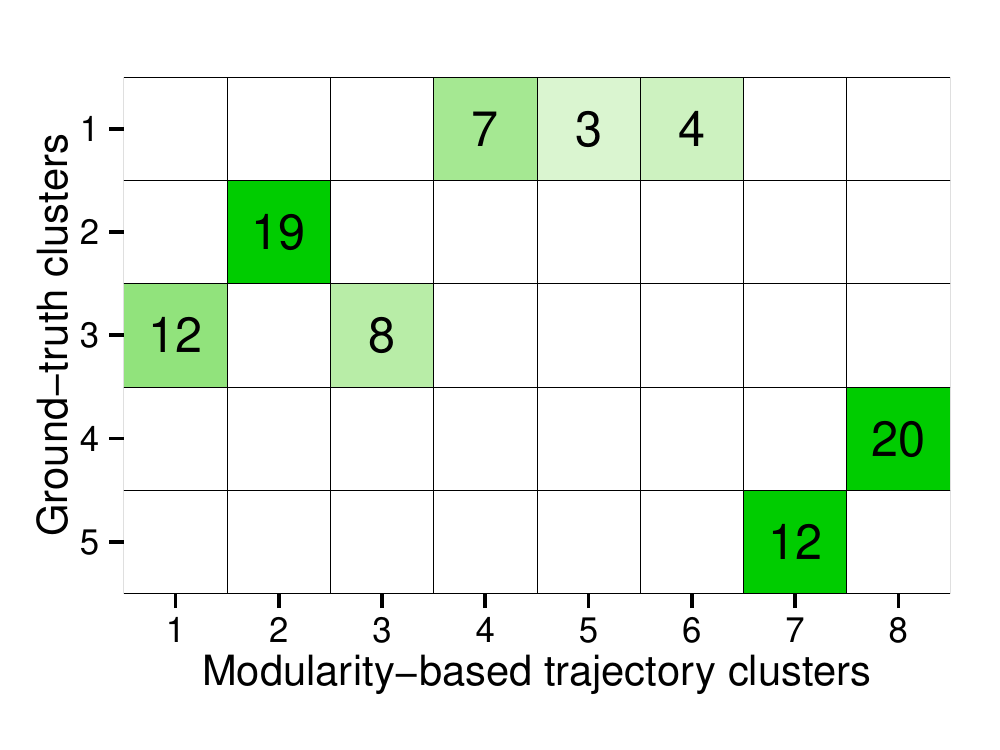}
		\label{fig:confusionmatrixmodularity}
	}
	\subfigure[Co-Clustering]{
		\includegraphics[width=0.48\textwidth, keepaspectratio]{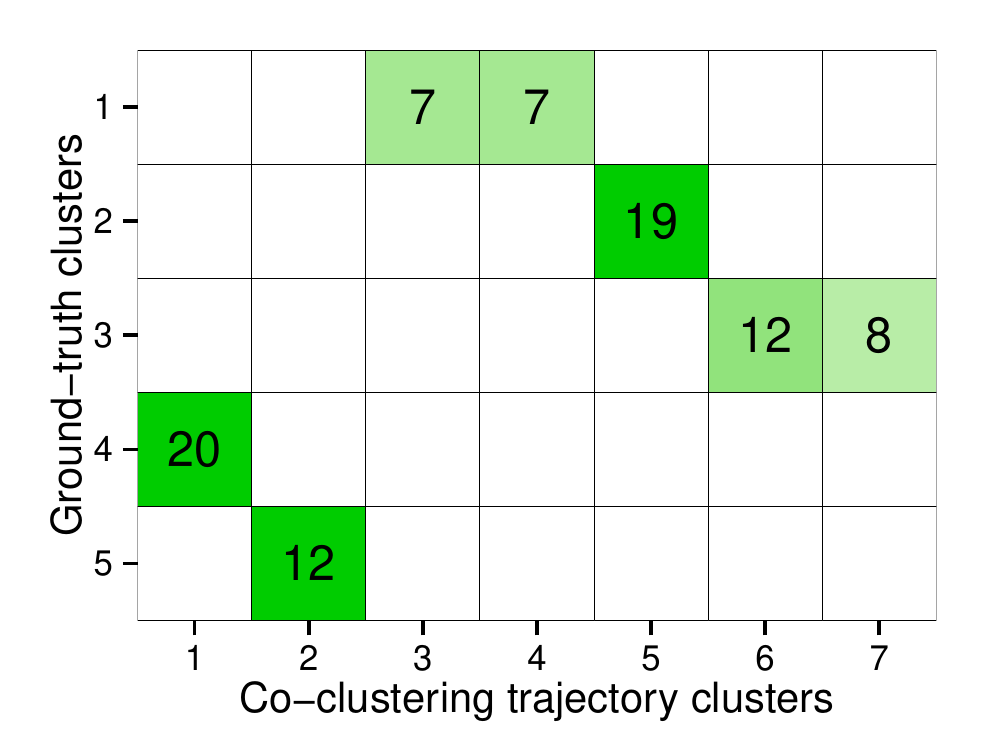}
		\label{fig:confusionmatrixcoclust}
	}
\caption{Confusion matrices of the original classes (ground-truth clusters) in the data and the clusters discovered by applying (a) the modularity-based approach and (b) the MODL co-clustering approach (cells are color-coded based on the ratio of the ground-truth cluster they contain). The modularity-based approach yields an Adjusted Rand Index (ARI) of 0.849, whereas MODL yields a slightly higher (i.e. better) ARI value of 0.862.}
\label{fig:confusionmatrices}
\end{figure}

Since the algorithm we apply is a hierarchical algorithm that produces a multi-level hierarchy of nested clusters, we can further refine the clusters in a given level by exploring their subsequent clusters. In the case of the dataset at hand, the second level reveals the existence of eight trajectory clusters. The confusion matrix between these clusters and the original classes is illustrated in Fig.~\ref{fig:confusionmatrixmodularity}. In this level, all the clusters contained in the partition are pure. Three of the original five classes were retrieved flawlessly whereas the two remaining clusters were further refined (class 1 was divided into three clusters and class 3 into two). This ``over-partitioning'' is legitimate and can be justified considering the variability of the trajectories contained in the concerned classes.

Co-clustering, on the other hand, directly retrieves a partition that is faithful to the original data (cf. Fig.~\ref{fig:confusionmatrixcoclust}). Here again two original classes were over-partitioned and three classes were retrieved correctly. Since the results (w.r.t. trajectory clusters) of applying MODL directly to the bipartite graph resemble those obtained by modularity-based clustering of the projected trajectory graph, it is more logical to use the former since it contains no preprocessing contrary to the latter where similarity calculations need to be performed first.

\subsection{Mutual Analysis of Trajectory and Segment Clusters}
\label{sec:MutualClusterAnalysis}

\begin{figure}[ht]
\centering
	\subfigure[Modularity]{\label{fig:Adjproj}\includegraphics[width=0.47\textwidth, keepaspectratio]{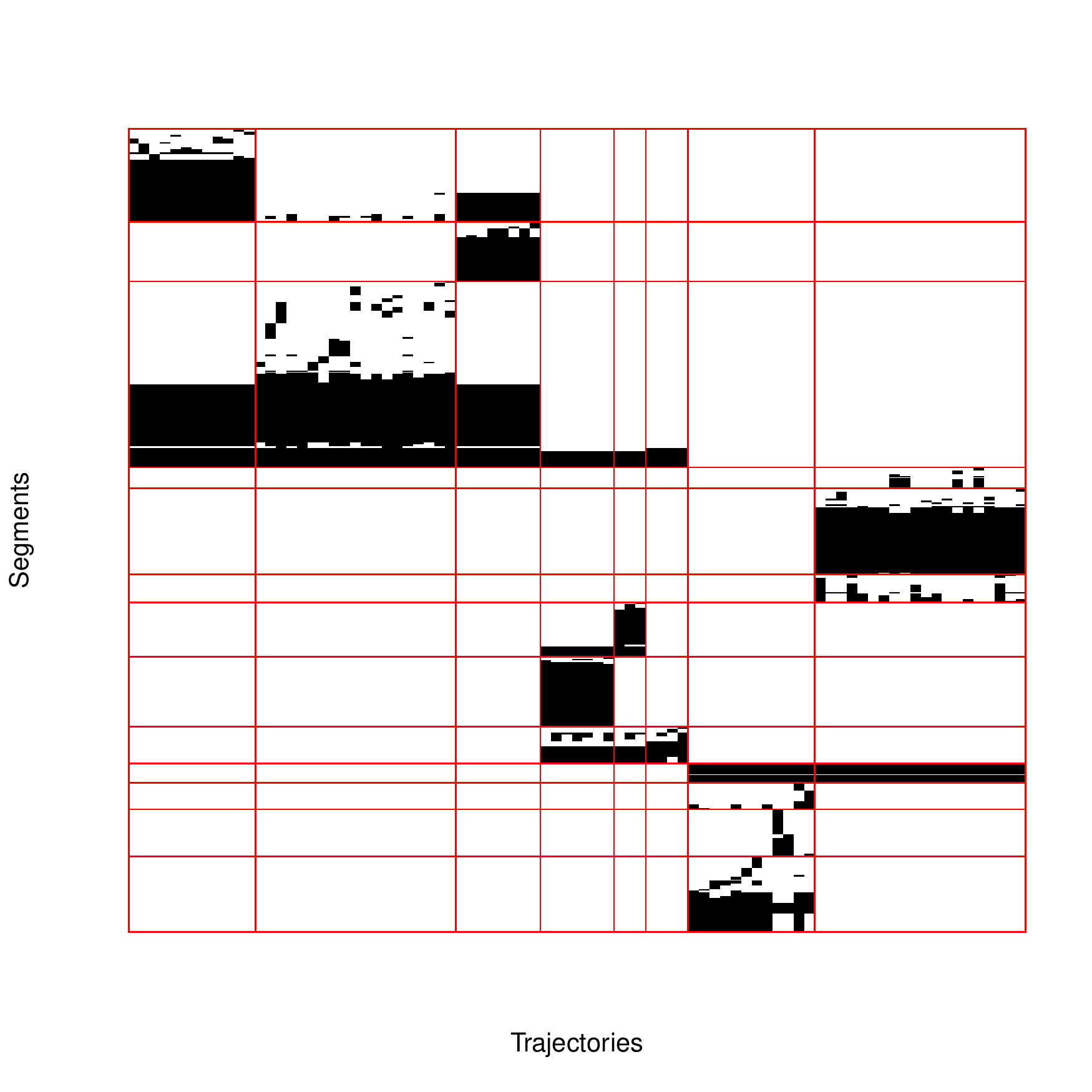}}
	\qquad
	\subfigure[Co-clustering]{\label{fig:Adjbi}\includegraphics[width=0.47\textwidth, keepaspectratio]{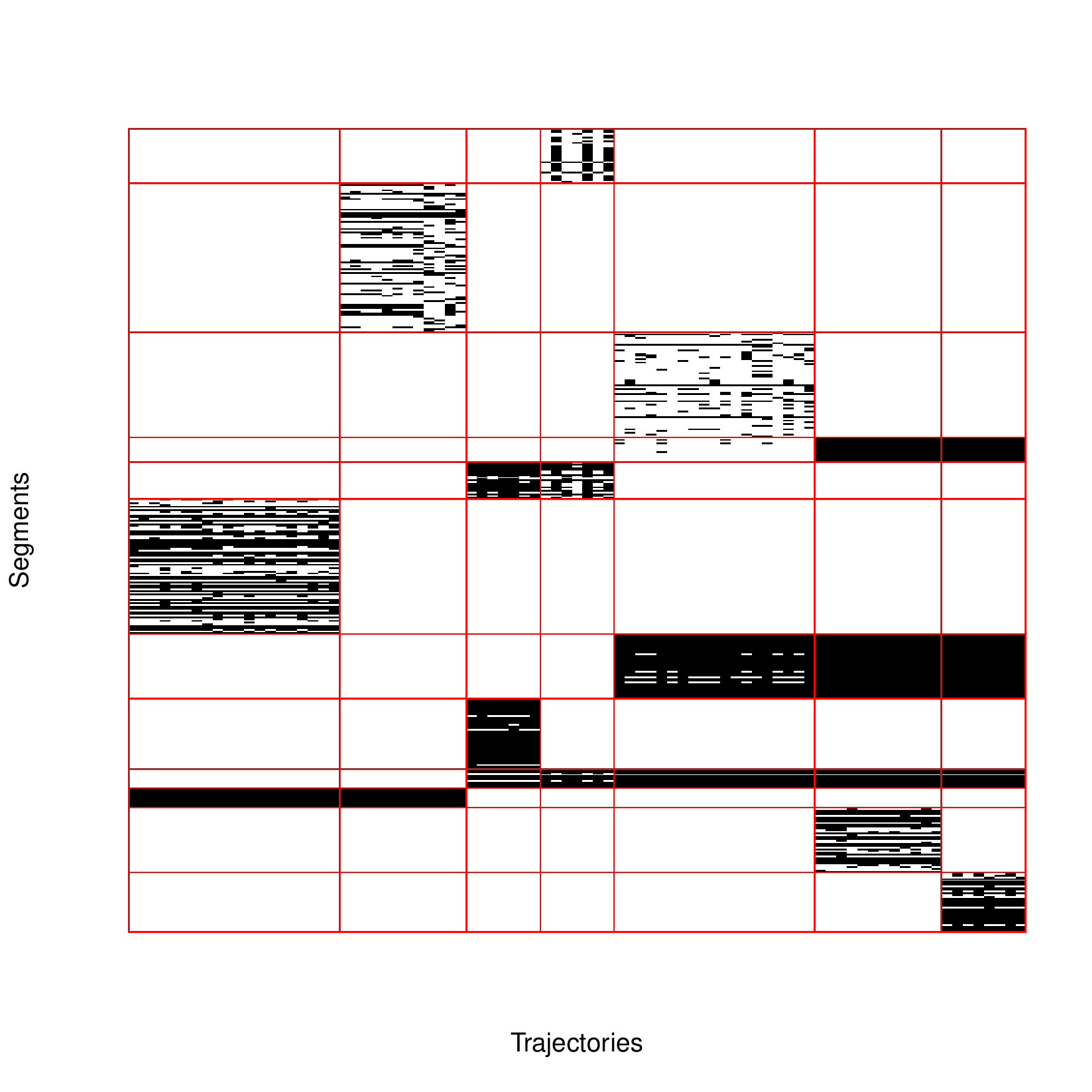}}
	\caption{Crossed matrices of trajectory clusters (columns) and road segment clusters (rows) retrieved through (a) modularity-based clustering and (b) co-clustering.}
\label{fig:matAdj}
\end{figure}

Let's now study the adjacency matrix of the original bipartite graph $\mathcal{G}$. We re-ordered the rows and columns of the matrix in order to bring together trajectories and segments belonging to the same clusters (cf. Fig.~\ref{fig:matAdj}). We observe in the case of the projected graphs (Fig.~\ref{fig:Adjproj}) that road segments are regrouped together based on common trajectories without accounting for the traffic's volume. Therefore, road segments that are rarely visited can be attached to segments that are visited frequently. This translates, when looking at the adjacency matrix, into the presence of blocs with  heterogeneous distributions in which some segments are travelled by all the trajectories in the cluster whereas others are only visited by a limited subset of trajectories. In co-clustering, on the other hand, segments are correlated based on usage which results in blocs of homogeneous densities (Fig.~\ref{fig:Adjbi}).

\begin{figure}[ht]
\centering
\subfigure[A highway hub]{\label{fig:hub}\includegraphics[width=0.4\textwidth, keepaspectratio]{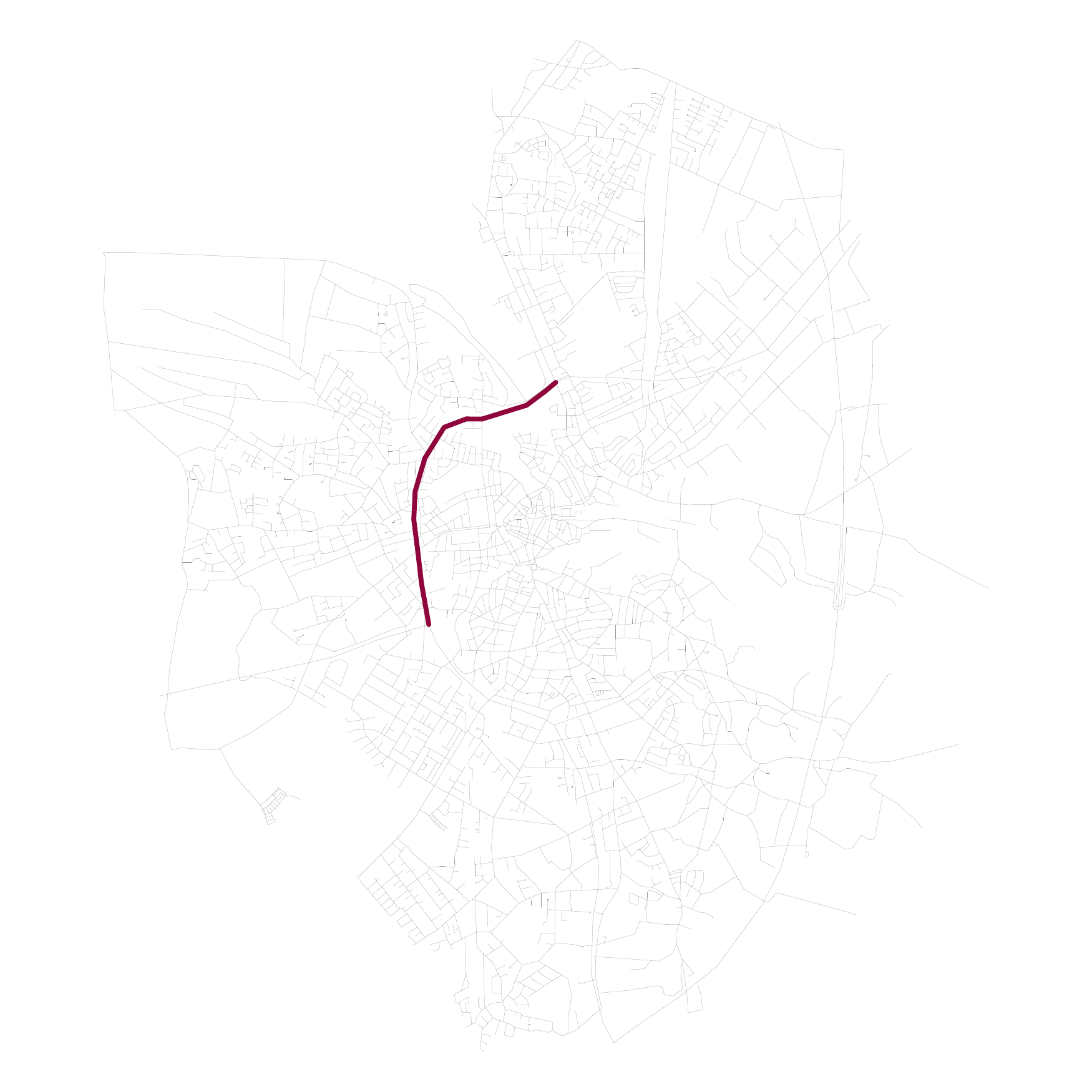}}
	\qquad
	\subfigure[Secondary roads leading to peripheral areas of the city]{\label{fig:secondaire}\includegraphics[width=0.4\textwidth, keepaspectratio]{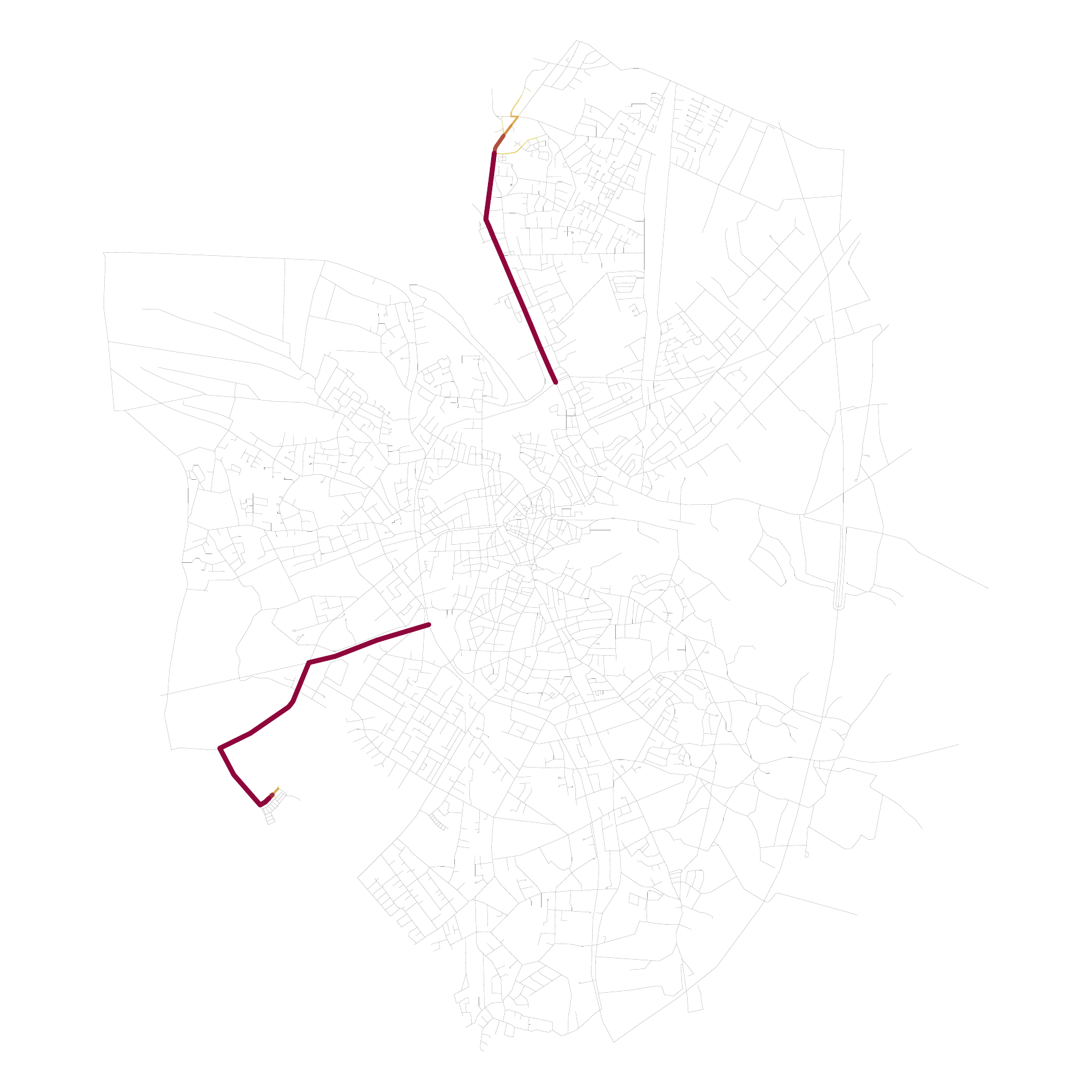}}
\caption{Example of road segment clusters.}
\label{fig:expBi1}
\end{figure}

By inspecting trajectory clusters and road segment clusters simultaneously, it is possible to characterize road segments based on the roles they play in traffic.This makes it possible to identify hubs that are frequently travelled by multiple groups of vehicles transiting to different regions (Figure~\ref{fig:hub}), secondary roads (Figure~\ref{fig:secondaire}), and even rarely frequented alleys. Therefore, our methodology makes it possible to characterize the topological structure of the underlying road network based on trajectories contributing their usage information.

Mutual information is frequently-used in co-clustering to quantify the correlations between partitions of the studied variables. These are, in our case, trajectories and road segments. Mutual information is always positive. High values of this metric usually indicate that trajectory clusters visit rather exclusive and unique segment clusters. We  use mutual information in our study to quantify the relationship between pairs of trajectory and segment clusters and their contribution to the model's mutual information. Given a cluster of trajectories $c_T$ and a cluster of road segments $c_S$, the contribution of the pair to mutual information, denoted $mi(c_S, c_T)$, is calculated as follows (\ref{eq:mi}).

\begin{equation}
mi(c_S,c_T)=P(c_S,c_T) \log \dfrac{P(c_S,c_T)}{P(c_S)P(c_T)}~.
\label{eq:mi}\end{equation}

Where $P(c_S,c_T)$ is the probability of a segment traversal to belong to a trajectory in $c_T$ and covering a road segment that belongs to $c_S$, $P(c_S)$ is the probability of visiting a segment belonging to $c_S$, and $P(c_T)$ is the probability of having a trajectory belonging to $c_T$.

A positive contribution to mutual information indicates that the number of visits of trajectories in $c_T$ to road segments in $c_S$ is higher than what is expected in case the two clusters were completely independent one from the other. Vice versa, a negative contribution is an indicator that quantity of traffic w.r.t. is inferior to normal. Finally, a null contribution to mutual information indicate that traffic either conforms to what is expected or is very low.

\begin{figure}[ht]
\centering
	\subfigure[Frequency]{\label{fig:freq} \includegraphics[width=0.45\textwidth, keepaspectratio]{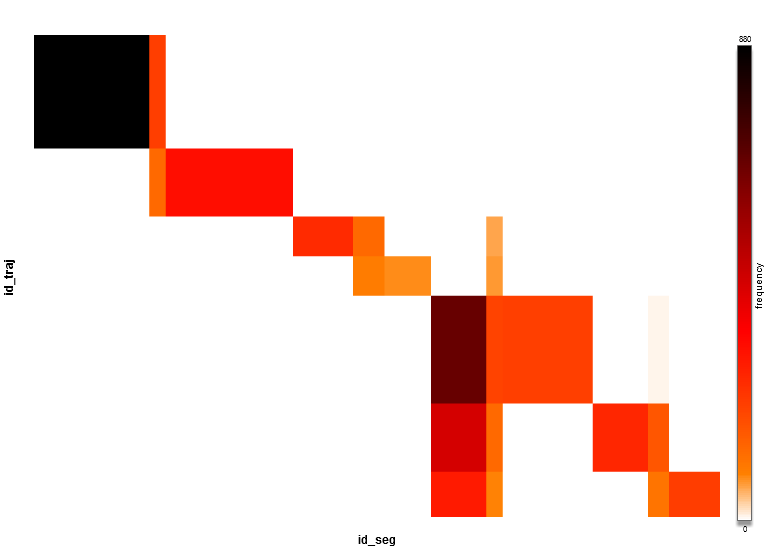}}
	\qquad
	\subfigure[Mutual information]{\label{fig:mi}\includegraphics[width=0.45\textwidth, keepaspectratio]{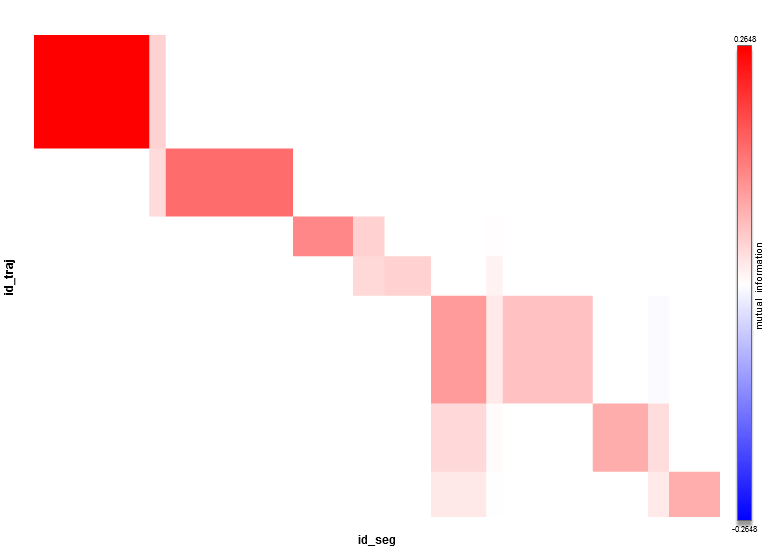}}
	\caption{Frequency and mutual information of the retrieved co-clusters.}
\label{fig:biInfo}
\end{figure}

Figure~\ref{fig:mi} presents the contribution to mutual information for each couple of co-clusters discovered in the dataset at hand. For instance, if we take the left, top-most co-cluster we can notice that the segments are exclusively travelled by members of a single trajectory cluster and that, vice versa, trajectories in this trajectory cluster travel almost uniquely on the members of this segment cluster. In this case, the trajectory cluster comprises $21.6\%$ of the studied trajectories and the road segments cluster $17.3\%$ of segments in the dataset. If we suppose that both clusters are independent, then we can 
expect no more than observing $21,6\% \times 17.3\% = 3.7\%$ of the total road segment traversals to be originating from both clusters. Here, however, we observe that no less than $17.3\%$ from the total traversals belong to this co-cluster which largely exceeds the expected traffic in case of unrelated and independent clusters. 

Notice that the mutual information contributes an information that is different from the frequency matrix. We can observe that some road segment clusters are significantly traversed by members belonging to multiple trajectory clusters. This behavior is quite characteristic of hubs that vehicles coming from different regions cross in order to attend different destinations. Some of these clusters have very small contrast w.r.t. mutual information which indicates that traffic on the hub is rather balanced.

\section{Related Work}
\label{sec:RelatedWork}
Approaches to trajectory clustering are mainly adaptations of existing algorithms to the case of trajectories. These include moving clusters \cite{Li_2004}, flock patterns \cite{Benkert_2006}, convoy patterns \cite{Jeung_2008a}, the TRACLUS partition-and-group framework \cite{Lee_2007}, etc. The aforementioned algorithms use euclidean-based similarities and distances and disregard the presence of an underlying network. Therefore, they can be used only in the case of unconstrained trajectories. The insightful idea of using a graph-based approach to cluster trajectory data was first introduced in \cite{Guo_2010}. The approach is applied to free moving trajectories and considers the latter as sets of GPS points. Unlike  our graph-based approaches, the authors do not rely on an underlying network as the basis of similarity calculations.

The first attempt to study the similarity between network-constrained trajectories is reported in \cite{Hwang_2005}. The proposition requiers a priori knowledge of points of interest in the road network and cannot, consequently, be used in an unsupervised learning context. An extension of moving clusters to network-constrained trajectories is presented in \cite{Liu_2008}. Roh et Hwang \cite{Roh_2010} present a network-aware approach to clustering trajectories where the distance between trajectories in the road network is measured using shortest path calculations. A baseline algorithm, using agglomerative hierarchical clustering, as well as a more efficient algorithm, called NNCluster, are presented for the purpose of regrouping the network constrained trajectories. In \cite{Kharrat_2008}, the authors describe an approach to discovering ``dense paths'' or sequences of frequently traveled segments in a road network. The approach is extended in \cite{Kharrat_2009} to study the temporal evolution of dense paths.

Our approaches differ from existing propositions on two key aspects. First, the majority of existing work use density-based algorithms that require fine-tuning of their parameter values and assume that trajectories in the same cluster have a rather homogeneous density (which is rarely the case as discussed in \cite{Roh_2010}). In contrast, we opt for non-parametric algorithms that rely on robust and well defined clustering quality criteria. Secondly, existing approaches often use flat clustering, thus producing a unique level of clusters that can be overwhelming to analyse in the case of large datasets. Our propositions produce hierarchies of nested clusters that are suitable for multi-level exploration: the user can start with a small number of clusters to quickly understand the macro-organization of flow dynamics in the road network, then proceed to refining clusters of interest to reveal more details.

\section{Conclusions}
\label{sec:Conclusions}

In this paper, we studied clustering network-constrained trajectory data from the angle of a bipartite graph clustering problem. We notably studied this problem from two different perspectives. At first, we considered the problem as a community detection problem conducted separately on two simple graphs (one depicting resemblances between trajectories and the other depicting similarities between road segments). Then we proceeded to co-cluster the bipartite graph directly to automatically retrieve clusters of interacting trajectories and road segments.

The main contribution of this work resides in its methodology of applying graph-based techniques to trajectory data in order to extract clusters describing mobility patterns in road networks. These clusters can be used by experts and road planners in conjunction with other data sources in order to understand trafic and driver behaviors. The applied clustering algorithms (modularity-based clustering used on the projected graphs and MODL used for co-clustering the bipartite graph) were essentially used to showcase and illustrate the interestingness of our problem formulation. As such, they can be replaced by other graph clustering and co-clustering approaches.

In future work, we will mainly focus on experimenting with the co-clustering technique on a larger scale (bigger datasets) as well as in the presence of noisy data.

 \bibliographystyle{apalike}
 \bibliography{bibliography.bib} 

\end{document}